# Texture Image Synthesis Using Spatial GAN Based on Vision Transformers

Elahe Salari[1], Zohreh Azimifar[2]
[12]Department of Computer Science and Engineering and Information Technology, Shiraz University, Shiraz, Iran;
[1]elahesalari@hafez.shirazu.ac.ir, [2]azimifar@cse.shirazu.ac.ir

***Abstract*— Texture synthesis is a fundamental task in computer vision, whose goal is to generate visually realistic and structurally coherent textures for a wide range of applications, from graphics to scientific simulations. While traditional methods like tiling and patch-based techniques often struggle with complex textures, recent advancements in deep learning have transformed this field. In this paper, we propose ViT-SGAN, a new hybrid model that fuses Vision Transformers (ViTs) with a Spatial Generative Adversarial Network (SGAN) to address the limitations of previous methods. By incorporating specialized texture descriptors such as mean-variance (μ, σ) and textons into the self-attention mechanism of ViTs, our model achieves superior texture synthesis. This approach enhances the model's capacity to capture complex spatial dependencies, leading to improved texture quality that is superior to state-of-the-art models, especially for regular and irregular textures. Comparison experiments with metrics such as FID, IS, SSIM, and LPIPS demonstrate the substantial improvement of ViT-SGAN, which underlines its efficiency in generating diverse realistic textures.***

## I. Introduction

Texture in image processing refers to the visual patterns and variations of pixel intensities within an image, representing spatial arrangements of fine details such as roughness and smoothness [1]. A crucial component in texture analysis is the texel, or texture element, which captures the smallest discernible unit of the texture pattern [2], playing a key role in applications like texture classification, segmentation, and synthesis [3]. Texture synthesis is a fundamental area of research in computer vision, aimed at generating visually appealing and realistic textures for various applications [4].

Over the years, texture synthesis methods have significantly evolved, from early tiling-based techniques to the more recent deep learning and GAN-based approaches. Tiling-based methods, one of the earliest techniques, involve replicating small texture patches to fill larger areas seamlessly [5], while stochastic-based methods generate textures by randomly selecting color values for each pixel, suitable for random textures [6], though they struggle with structured textures [7].

Patch-based methods emerged as a more advanced approach, replicating local texture patterns by stitching together small patches from the input texture, producing more coherent textures [8]. In contrast, example-based methods leverage databases of exemplar textures, finding similar patches and assembling them to synthesize new textures [9], with algorithms like Image Analogies [10] and Markov Random Fields [11] being prominent examples.

Recent advancements include deep learning-based methods, which have revolutionized texture synthesis using generative models like Convolutional Neural Networks (CNNs) [12], Generative Adversarial Networks (GANs) [13], and Vision Transformers (ViTs). CNN-based approaches generate textures using learned representations, while GANs employ an adversarial framework where a generator creates realistic textures, and a discriminator attempts to distinguish between real and synthesized ones [14]. ViTs process images as sequences of patches, leveraging attention mechanisms to capture complex relationships within textures, enhancing texture synthesis capabilities.

Texture synthesis methods rely on feature extraction techniques, such as mean and variance $(\mu, \sigma)$ and textons. The $(\mu, \sigma)$ together assess the average pixel intensity and its variation, reflecting the texture's brightness, contrast, and roughness [15]. Meanwhile, textons represent the basic perceptual elements that characterize textures, serving as repetitive structural units that define their patterns [16]. Overall, this paper examines two models, GANs and Vision Transformers, for texture synthesis and presents our work on combining these two architectures, leveraging the feature extractions of $(\mu, \sigma)$, along with textons, to propose a novel structure [13].

## II. Related Work

The history of texture synthesis methods dates back several decades, with early techniques focusing on the statistical properties of textures. These methods aimed to capture the essential features of textures, leading to the development of algorithms that utilized co-occurrence matrices and filters to analyze and synthesize textures. Among the first statistical models, those proposed by Haralick et al. [17] in the 1970s laid the foundation for modern approaches to the analysis and classification of textures.

Statistical models for texture synthesis have evolved significantly, utilizing various probabilistic frameworks. One popular approach involves using Markov Random Fields (MRFs) [18], which model textures as a collection of random variables that exhibit spatial dependencies. These methods enable the generation of textures by sampling from a learned probability distribution, allowing for coherent synthesis while preserving local and global structure. Such models have paved the way for more advanced techniques that leverage the underlying statistical properties of textures for effective synthesis.

In the realm of neural networks, Convolutional Neural Networks (CNNs) were among the first to be applied to texture synthesis. Gatys et al. [12] proposed a groundbreaking approach, which introduced a method for generating textures by utilizing the hierarchical feature representations learned by CNNs. This approach demonstrated the potential of CNNs to capture intricate texture patterns and enabled the synthesis of high-quality textures. Variational Autoencoders (VAEs) have also been employed for texture synthesis, where the authors utilized recurrent architectures to generate diverse and coherent textures by leveraging learned latent representations [19].

Generative Adversarial Networks (GANs) have transformed texture synthesis by introducing an adversarial framework for generating realistic textures. Jetchev et al. [20] highlights the effectiveness of GANs in capturing complex texture distributions through the interplay between a generator and a discriminator, resulting in high-quality synthesized textures that mimic real-world patterns.

Recently, Vision Transformers have emerged as a promising approach for texture synthesis. Lu [21] demonstrates how Transformers can effectively synthesize textures by treating images as sequences of patches, allowing for the capture of long-range dependencies and enhancing the quality of generated textures.

## III. METHODOLOGY

In this section, we present our proposed method for texture synthesis, which integrates Transformer blocks within the GAN architecture, replacing traditional convolutional layers [22]. This hybrid approach leverages the long-range dependency modeling of Transformers while maintaining the adversarial framework of GANs. For texture synthesis, we utilize two texture descriptors, namely mean-variance $(\mu, \sigma)$ and textons. Additionally, we incorporate the loss function defined for texture in the Spatial GAN model [20] into our hybrid architecture. We will briefly introduce the background of Vision Transformers and then present the details of our primary model, ViT-SGAN.

### *A. Vision Transformers Background*

Vision Transformers (ViTs) [23] revolutionize image processing, mainly in the field of image classification. Originally developed for natural language processing, ViTs split images into fixed-size patches and treat them as tokens in language models. The self-attention mechanism [24] of ViTs encodes spatial relations between patches, which turn out very successful at capturing and generating the texture pattern of the whole image with high accuracy.

At the heart of Vision Transformers is the self-attention mechanism, which computes the relevance of one patch to others in the image sequence. Given a sequence of patches $X \in \mathbb{R}^{n \times d}$, where $n$ is the number of patches and $d$ is the embedding dimension. The input sequence $X$ is first projected onto three learnable weight matrices for Queries $(W_Q)$, Keys $(W_K)$, and Values $(W_V)$ to get $Q = XW_Q$, $K = XW_K$, and $V = XW_V$. The output $Z$ of the self-attention layer is computed as:

$$Z = softmax\left(\frac{QK^T}{\sqrt{d_q}}\right)V \quad (1)$$

where the dot-product of the query with all keys is normalized using the softmax operator to obtain attention scores. Each entity becomes a weighted sum of all entities in the sequence, where the weights are determined by these attention scores. This global context aggregation enables the Vision Transformer to effectively capture complex texture relationships that are essential for high-quality texture synthesis.

### *B. The ViT-SGAN model*

We combine the Vision Transformer (ViT) architecture with Generative Adversarial Networks (GANs) for texture image generation. Inspired by Lee et al. [22] on studying the combination of Transformers with GANs for generating images, we handle instability in standard GAN regularization methods when combined with self-attention by introducing texture-specific regularization techniques to make this training process stable. In our approach, the convolutional layers in both the Generator and Discriminator are replaced with Transformer-based layers Figure 1, which optimizes the architecture specifically for generating high-quality texture images. The Generator structure follows the design in [10], where the Transformer-based generator is tailored to efficiently synthesize texture patterns and enhance the overall quality of generated images.

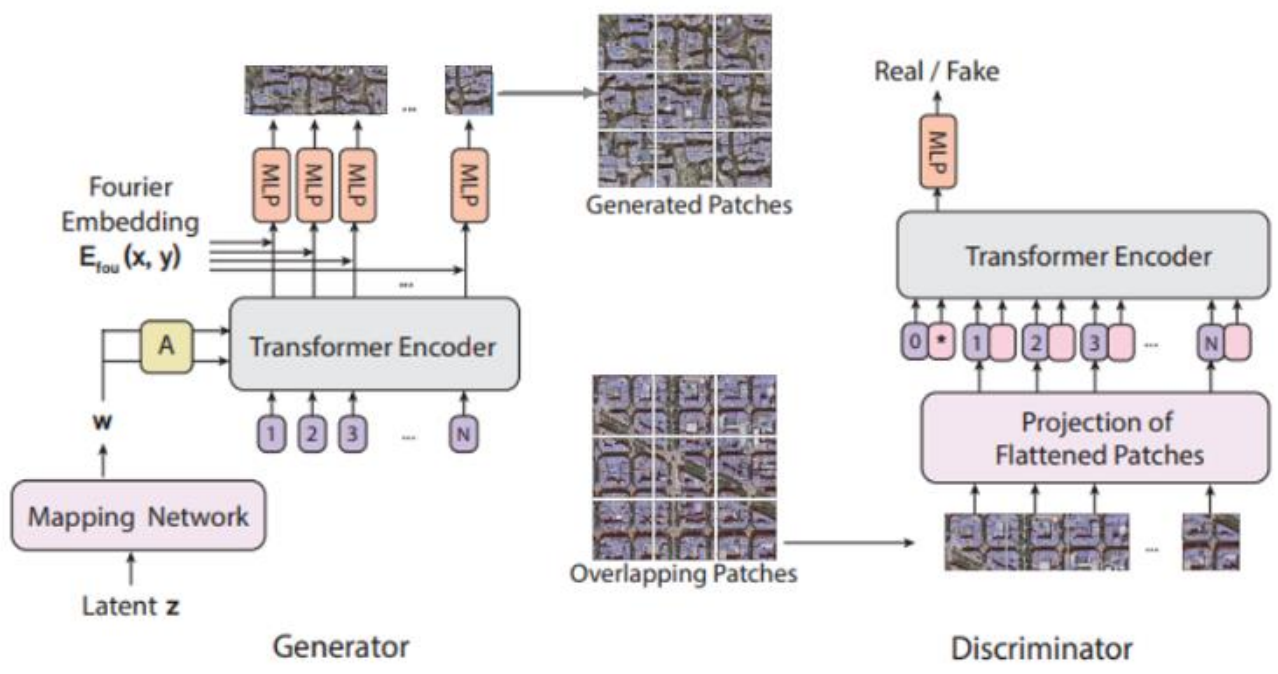

Figure 1: ViT-GAN Architecture [22]: The generator and discriminator are both built using Transformer blocks.

To further enhance the performance of the Vision Transformer (ViT) in texture image generation, we focus on refining the core component: the Self-Attention layer. This layer is modified specifically for texture images by adding a texture-based regularizer inside the Transformer in the GAN's discriminator. The adjustments are tailored to capture texture-specific patterns, utilizing descriptors such as $(\mu, \sigma)$, and Textons. These descriptors are incorporated in the Self-Attention mechanism to fine-tune the model for effective high-quality texture generation and analysis.

#### *1) $(\mu, \sigma)$ texture descriptor*

In our work, we adopt Lu's texture descriptor definition [21], as it provides a fast, memory-efficient method for processing textures, ideal for modifying the Transformer blocks in discriminator for texture synthesis. The core idea is to compute the mean (μ) and variance (σ) of pixel values within each image patch, with these values being incorporated into the Transformer blocks to specialize the Self-Attention mechanism for texture analysis. For each patch, the mean is calculated by first computing the mean of each row (height) and column (width) across all color channels, followed by matrix multiplication to produce a mean matrix that retains the patch's dimensions, which is subsequently used in the Attention formula.

Similarly, the variance for each patch is calculated by determining the deviation from the mean for both rows and columns, as expressed in the following formulas:

$$\mu = \frac{1}{M}\sum_{k=1}^{M} x_k \cdot \frac{1}{N}\sum_{k=1}^{N} x_k \quad (2)$$

$$\sigma^2 = \frac{1}{M}\sum_{k=1}^{M}(x_k - \mu_{row})^2 \cdot \frac{1}{N}\sum_{k=1}^{N}(x_k - \mu_{col})^2 \quad (3)$$

where $M$ defines the number of rows and $N$ defines the number of columns of each image patch. These texture descriptors $(\mu, \sigma)$ are then incorporated into the three weighted matrices ( $W_{Query}$ , $W_{Key}$ , $W_{Value}$ ) within each Transformer block, which are computed by multiplying the weight matrices with the mean and variance matrices. The mean and variance calculations differ between real images $(\mu, \sigma)$ and generated images $(\hat{\mu}, \hat{\sigma})$, with these metrics forming the basis for comparing real and generated texture features in the discriminator. Specifically, the Query (Q), Key (K), and Value (V) matrices are adjusted as follows:

$$Q = W_q\left((\hat{\mu} - \mu)^2 + (\widehat{\sigma^2} - \sigma^2)\right)^{\frac{1}{2}} \quad (4)$$

$$K = W_k\left((\hat{\mu} - \mu)^2 + (\widehat{\sigma^2} - \sigma^2)\right)^{\frac{1}{2}} \quad (5)$$

$$V = W_v\left((\hat{\mu} - \mu)^2 + (\widehat{\sigma^2} - \sigma^2)\right)^{\frac{1}{2}} \quad (6)$$

These formulas represent how the differences between the real $(\mu, \sigma)$ and generated $(\hat{\mu}, \hat{\sigma})$ texture descriptors are incorporated into the self-attention mechanism to enhance texture discrimination within the GAN's Transformer-based architecture.

Incorporating these descriptors into the Attention mechanism for the discriminator requires adjustments to the Attention formula. Specifically, the dot product between Query (Q) and Key (K) is replaced with the Euclidean (L2) distance, ensuring Lipschitz continuity, which is crucial for the stability of GANs [22]. The modified Attention formula is shown below:

$$Attention_h(X) = softmax\left(\frac{d(Q,K)}{\sqrt{d_h}}\right)V \quad (7)$$

wherein $d(.,.)$ represents the L2 distance between matrices. This modification helps the discriminator distinguish subtle texture patterns more effectively, and to avoid overfitting to local texture details, a 10-pixel overlap is introduced between patches, which enhances the model's local perception by encouraging interaction between neighboring patches. This adjustment improves the generator's ability to synthesize more coherent and globally consistent textures.

### 2) Texton Texture Descriptor

The Texton texture descriptor method extracts texture features through edge orientation and color feature detection, followed by Texton identification. Edge orientation features are detected using the Sobel operator, generating 18 orientation bins, as described in previous studies [25]. Color features are extracted by quantizing the RGB components into 64 bins, which are then combined with edge data. Texton detection is performed on a 2x2 grid to enhance texture discrimination, identifying common texture patterns, or Textons (T1 to T4), where two pixels are likely to have similar values based on color or edge orientation [26]. This multi-Texton histogram (MTH) method is visualized in Figure 2 [27].

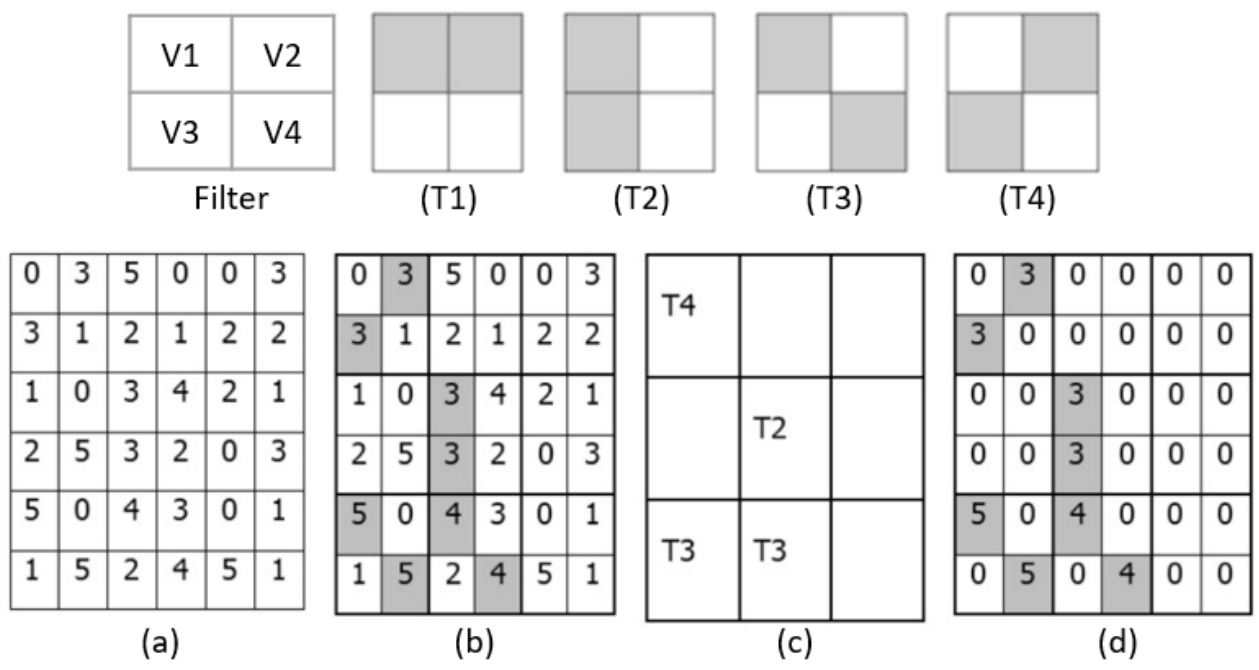

Figure 2: Model schematic of Texton detection in MTH. (a) Original image containing edge and color information; (b) Four identified Texton types in a 2x2 grid; (c) Texton type/name; (d) Feature extraction result through Texton detection [27].

Once the texture features are extracted, they are used in the Attention formula within the Vision Transformer (ViT) blocks. The extracted feature matrix is incorporated into the weighted matrices for the attention mechanism, as shown in Figure 3. The Attention formula is computed as:

$$Attention_h(X) = softmax\left(\frac{(W_q X)\cdot(W_k X)^T}{\sqrt{d_h}}\right)W_v X \quad (8)$$

where $X$ represents the texture features derived from the Texton method. This incorporation enhances texture recognition and analysis within the Transformer-based GAN architecture.

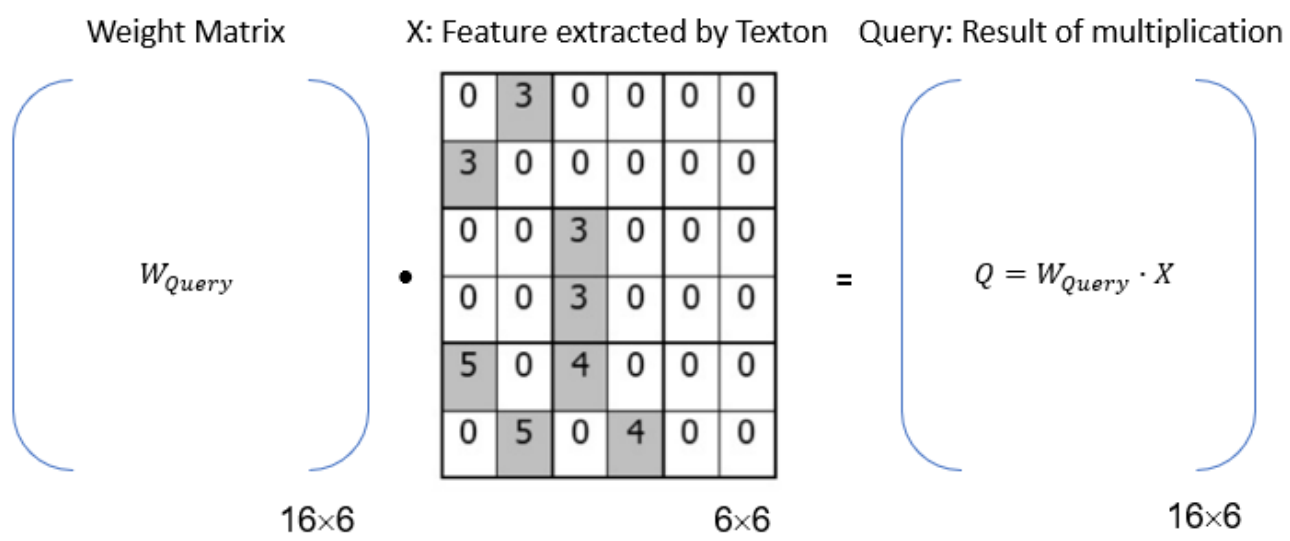

Figure 3: Example of matrix multiplication for Query calculation, where the weighted matrix $W_{Query}$ is dot-multiplied with the Texton feature matrix to produce the Query matrix. This process is repeated for the Key and Value matrices in the Attention formula.

In the ViT-SGAN model, we integrate the loss function from Jetchev's SGAN approach [20] into the ViTGAN architecture to enhance texture synthesis. The standard ViTGAN uses Binary Cross Entropy (BCE) loss for both the generator and discriminator. However, to better capture texture-specific features, we adopt the SGAN loss function proposed by Jetchev [20], which is designed for texture synthesis. This loss function guides the generator to focus on texture structure while assisting the discriminator in distinguishing between real and generated textures. The SGAN loss is expressed as:

$$min_G \, max_D V(D,G) = \frac{1}{lm}\sum_{\lambda=1}^{l}\sum_{\mu=1}^{m} E_{Z\sim p_z(z)}[\log(1-D_{\lambda\mu}(G(Z)))] + \frac{1}{lm}\sum_{l=1}^{l}\sum_{m=1}^{m} E_{X_0\sim p_{data}(X)}[\log(D_{\lambda\mu}(X_0))] \quad (9)$$

By incorporating this loss into the ViT-SGAN, we improve the ability of both the generator and discriminator to capture and distinguish texture features, leveraging the structure proposed by Jetchev for enhanced texture synthesis.

## IV. Experimental Setup

We follow the training implementation parameters outlined in [22], as shown in Table 1. The model was initially trained with 4 Attention Head modules using Adam optimization with parameters $\beta_1 = 0.0$, $\beta_2 = 0.99$, and a learning rate of 0.002. It was trained on datasets such as CIFAR-10, LSUN bedroom, and CelebA. To adapt the model to our specific texture data, we use the pre-trained model weights and fine-tune the model on texture samples. This fine-tuning process, which includes modifications to the Self-Attention architecture, is carried out over 5000 epochs. The fine-tuning was performed on an NVIDIA GeForce GTX 1080 graphics card with 16GB of memory. Due to hardware limitations, the number of Transformer blocks was reduced to one for this fine-tuning process.

TABLE I. Implementation parameters for model training [22]

| Parameters | Value |
|---|---|
| Resolution | 32 × 32 |
| Number of ViT Blocks | 4-blocks |
| Feature Dimension | 384 |
| Hidden Dimension | 1536 |
| Patch size | 4 × 4 |
| Sequence len | 64 |

## V. Results and Discussions

In this study, we implemented SGAN [20] and Texture-ViT [21] models on our texture data samples to establish a baseline for comparison with our proposed ViT-SGAN model. The ViT-SGAN incorporates modifications to the self-attention mechanism and the SGAN loss function, along with the integration of two texture-specific descriptors: (μ, σ) and Texton. We conducted a comparative analysis to evaluate the performance of these descriptors individually, as well as in relation to the original SGAN and Texture-ViT models. Due to the inclusion of Transformer blocks, enhancements in Transformer Attention, and the application of a texture-specific loss function, we hypothesized that ViT-SGAN would outperform the previous models. The model was trained with the specified parameters over 5000 iterations, and the results for the (μ, σ) and Texton descriptors, alongside the performance of SGAN and Texture-ViT, are shown in Figure 4.

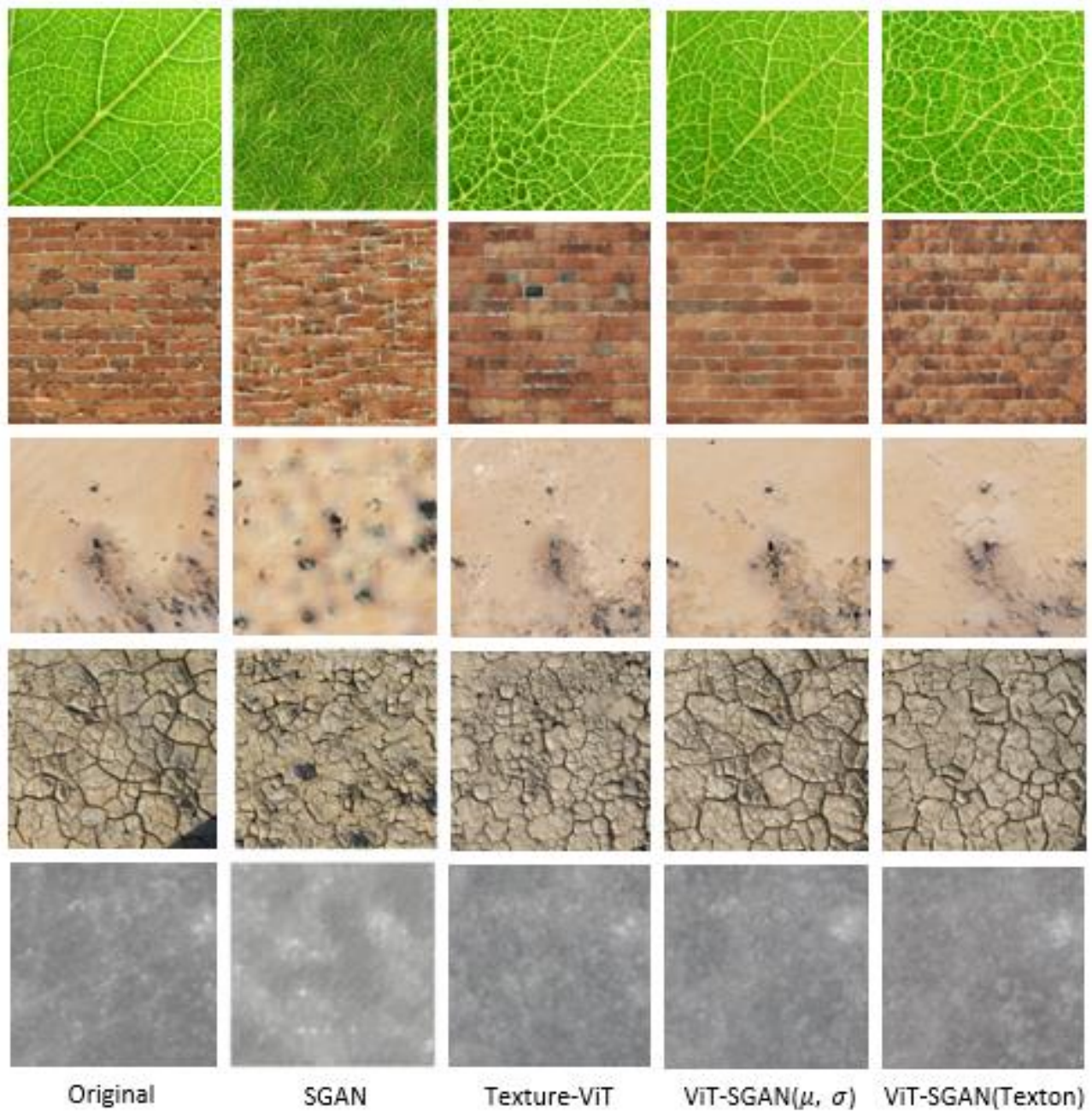

Figure 4: Results of the proposed ViT-SGAN model with two texture descriptors compared to SGAN and Texture-ViT models. The two rows of images above belong to the regular texture category, and the last three rows are categorized as irregular textures.

Textures are generally categorized into two types: regular and irregular textures. Regular textures exhibit a certain structure and pattern, while irregular textures are more random and lack a discernible order. In our study, we observed and classified our texture samples into these two categories. As seen in Figure 4, the first two rows of images belong to the regular texture category, while the following three are classified as irregular textures.

When comparing our proposed model to existing methods like SGAN and Texture-ViT, the comparison can be extended to these texture categories. For instance, SGAN and Texture-ViT models struggled to generate regular textures and failed to capture the underlying structure. However, these models performed slightly better in generating irregular textures, where the absence of a defined structure led to acceptable outputs despite the lack of order.

Furthermore, comparing both texture descriptors, very close results are obtained. Both perform well in generating irregular textures; however, the (μ, σ) descriptor slightly outperforms in generating regular textures because it preserves details and structure and produces images much closer to the original. Still, the Texton descriptor has shown good performance in generating irregular textures and acceptable results for regular textures.

In this model, we evaluated the results using standard metrics such as FID, IS, SSIM, and LPIPS. FID (Fréchet Inception Distance) measures the similarity between real and generated images, IS (Inception Score) assesses image quality and diversity, SSIM (Structural Similarity Index) evaluates structural similarity, and LPIPS (Learned Perceptual Image Patch Similarity) quantifies perceptual differences. The results for each metric, based on a sample of a regular texture, are shown in Table2 and for an irregular texture sample in Table 3.

TABLE II. EVALUATION METRICS FOR A REGULAR TEXTURE SAMPLE.

| | FID↓ | IS↑ | SSIM↑ | LPIPS↓ |
|---|---|---|---|---|
| SGAN | 125.6 | 1.22 | 0.1 | 0.573 |
| Texture-ViT | × | × | 0.29 | 0.385 |
| ViT-SGAN ($\mu$, $\sigma$) | **53.7** | **2.73** | **0.42** | **0.324** |
| ViT-SGAN (Texton) | 61.1 | 2.2 | 0.35 | 0.367 |

TABLE III. EVALUATION METRICS FOR A IRREGULAR TEXTURE SAMPLE.

| | FID↓ | IS↑ | SSIM↑ | LPIPS↓ |
|---|---|---|---|---|
| SGAN | 62.45 | 2.8 | 0.33 | 0.494 |
| Texture-ViT | × | × | 0.49 | 0.345 |
| ViT-SGAN ($\mu$, $\sigma$) | **14.3** | **3.95** | **0.66** | **0.309** |
| ViT-SGAN (Texton) | 25.1 | 3.24 | 0.55 | 0.331 |

## VI. CONCLUSION AND FUTURE WORK

The SGAN-ViT model has demonstrated superior results compared to existing models, both in terms of evaluation metrics and the structural quality of the generated textures. While this marks a significant improvement over previous methods, there is still quite a noticeable perceptual gap between the generated images and the original textures; hence, further improvements are still needed.

As future work, we would like to investigate the application of other texture descriptors. Many texture descriptors have been proposed over the years, and including those in our hybrid model may give even more refined and precise results. Another direction for improving texture synthesis is by using diffusion models, which are recently popular for their strong image generation capability. Our current work is aimed at coupling diffusion models, and we will report on these advancements in future publications.